\documentclass[letterpaper, 10 pt, conference]{ieeeconf}  

\IEEEoverridecommandlockouts                              
\usepackage{mathtools,amssymb,commath}
\usepackage{amsmath}
\usepackage{graphicx,import}
\usepackage{hyperref}
\hypersetup{colorlinks=true}

\usepackage{mathtools,amsthm}

\title{\LARGE \bf Modeling and Control of Humanoid Robots in Dynamic \\ Environments: iCub Balancing on a Seesaw
}

\author{Gabriele Nava, Daniele Pucci, Nuno Guedelha, Silvio Traversaro, \\ Francesco Romano, Stefano Dafarra and Francesco Nori$^{1}$
\thanks{*This paper was supported by the FP7 EU project CoDyCo (No. 600716 ICT 2011.2.1 Cognitive Systems and Robotics)}
\thanks{$^{1}$ All authors belong to the iCub Facility department, Istituto Italiano di Tecnologia,
        Via Morego 30, Genoa, Italy
        {\tt\small name.surname@iit.it}}%
}

\newtheorem{lemma}{\bf{Lemma}}

\DeclareMathOperator*{\argmin}{argmin}

\makeatletter
\newsavebox\myboxA
\newsavebox\myboxB
\newlength\mylenA
\newcommand*\xoverline[2][0.75]{%
    \sbox{\myboxA}{$\m@th#2$}%
    \setbox\myboxB\null
    \ht\myboxB=\ht\myboxA%
    \dp\myboxB=\dp\myboxA%
    \wd\myboxB=#1\wd\myboxA
    \sbox\myboxB{$\m@th\overline{\copy\myboxB}$}
    \setlength\mylenA{\the\wd\myboxA}
    \addtolength\mylenA{-\the\wd\myboxB}%
    \ifdim\wd\myboxB<\wd\myboxA%
       \rlap{\hskip 0.5\mylenA\usebox\myboxB}{\usebox\myboxA}%
    \else
        \hskip -0.5\mylenA\rlap{\usebox\myboxA}{\hskip 0.5\mylenA\usebox\myboxB}%
    \fi}
\makeatother

\begin{document}

\maketitle
\thispagestyle{empty}
\pagestyle{empty}

\begin{abstract}

Forthcoming applications concerning humanoid robots may involve physical interaction between the robot and a dynamic environment. In such scenario, classical balancing and walking controllers that neglect the environment dynamics may not be sufficient for achieving a stable robot behavior. This paper presents a modeling and control framework for balancing humanoid robots in contact with a dynamic environment. We first model the robot and environment dynamics, together with the contact constraints. Then, a control strategy for stabilizing the full system is proposed. Theoretical results are verified in simulation with robot iCub balancing on a seesaw.  

\end{abstract}

\section{Introduction}
\label{sec:introduction}

Prospective applications for robotics may require robots to step out on protected and well-known workspaces and physically interact with dynamic, human-centered environments. In this context, a humanoid robot is required to balance, walk, perform manipulation tasks and -- even more important -- safely interact with humans. 

The importance of controlling the robot interaction with the environment calls for the design of torque and impedance control algorithms, capable of exploiting the forces the robot exerts at contact locations, for performing balancing and walking tasks \cite{ott2011,Stephens2010,farnioli2015}. However, the applicability of such controllers in a real scenario is often limited by the assumption that the robot is in contact with a rigid, static environment. From the modeling point of view, this results in neglecting the environment dynamics, i.e. the robot is subject to a set of purely \emph{kinematic} constraints \cite{deluca1994}. This assumption may be a limitation in case the robot is walking on debris or uneven ground. Different solutions that make use of adaptive or robust controllers are available in literature \cite{Lee2012,Kim2007,morisawa2012,hyon2009}. There are also situations in which the environment dynamics may be known \emph{a priori}, e.g. interacting with a wheeled chair, balancing on a moving platform or even interacting with humans. This leads to the development of control strategies that try to stabilize both the robot and the contact dynamics \cite{deluca1994,anderson2010}.

\begin{figure}[ht!]
   \centering
   \includegraphics[scale=0.20]{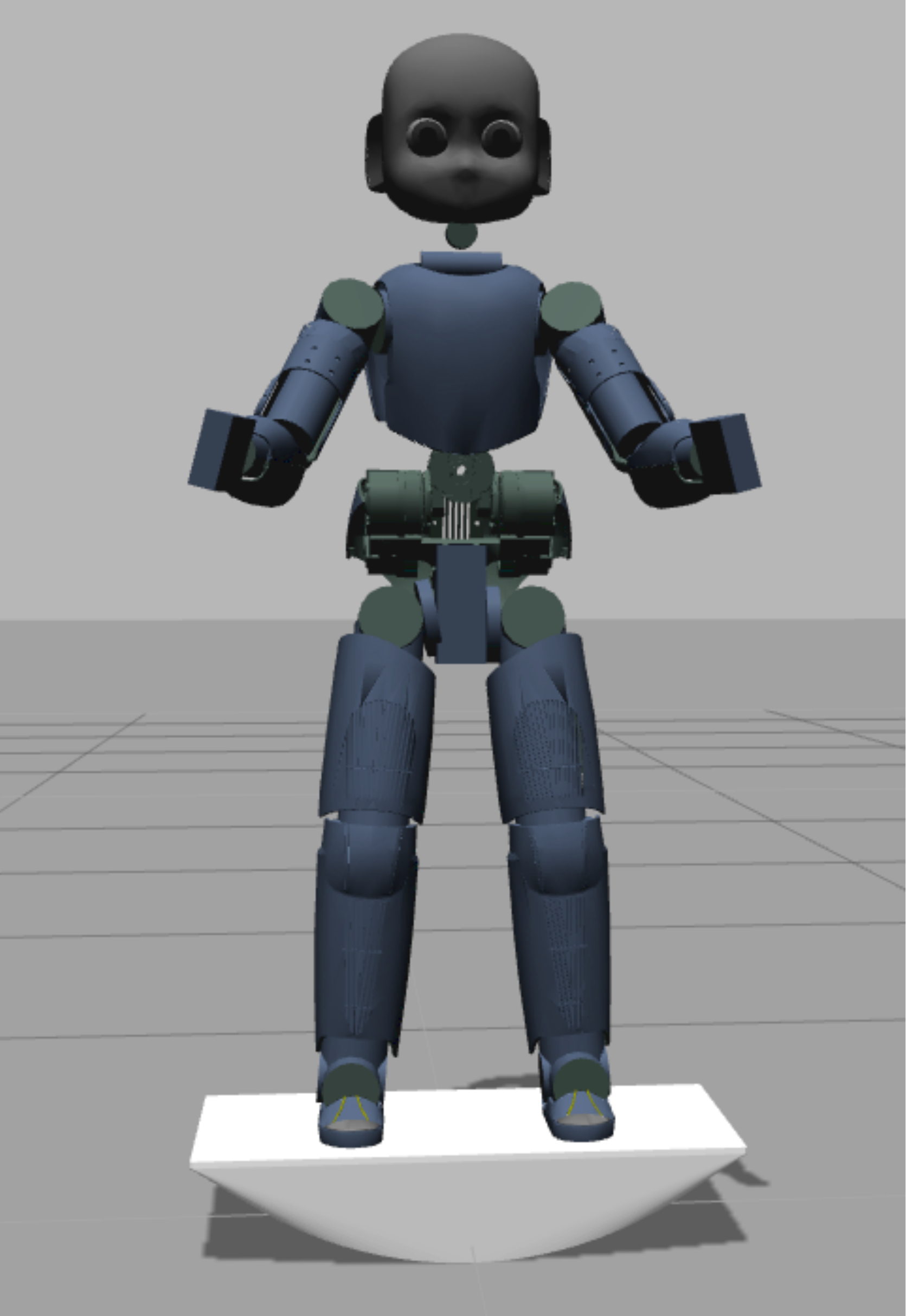}
   \caption{iCub balancing on a seesaw.}
   \label{fig:seesaw}
\end{figure}

On the modeling side, the fixed base assumption may be a strong limitation for a humanoid robot, capable  -- at least theoretically-- to move from place to place without being physically attached to the ground. The \textit{floating base} formalism ~\cite{Featherstone2007}, i.e. none of the robot links has an \emph{a priori} constant pose with respect to an inertial reference frame, is particularly well suited for modeling humanoid robots dynamics. However, the control problem is complicated by system's underactuation, that forbids full state feedback linearization~\cite{Acosta05}. 

At the control level, an efficient algorithm for balancing and walking of humanoid robots is the so-called 
\textit{momentum-based} control \cite{Lee2012}, which often exploits \textit{prioritized stack-of-tasks}. In particular, the primary control objective 
is the stabilization of \textit{centroidal momentum} dynamics~\cite{Orin2013}. Momentum control can be achieved by properly choosing the contact forces the robot exerts at contact locations \cite{Stephens2010,Herzog2014,Frontiers2015}. 
Robot joint torques are then used for generating the desired forces. To get rid of the (eventual) actuation redundancy associated with momentum control, a lower priority task is usually added during the stabilization of the robot momentum, whose main role is the stabilization of the so-called robot \textit{zero dynamics}~\cite{ISIDORI2013}. 

In this paper, we propose a modeling and control framework for balancing a humanoid robot on a seesaw board (Fig. \ref{fig:seesaw}). Similar problems have been already addressed in literature \cite{anderson2010,hyon2009}. In particular, in \cite{anderson2010} the authors developed a torque control strategy based on weighted control policies, together with online and offline model adaptations, for balancing a humanoid robot on a moving platform. We follow a similar approach, but we then design a different control algorithm. In particular, the aim of our paper is to apply a momentum-based control strategy in case of balancing in a dynamic environment.

The remaining of the paper is organized as follows. Section \ref{sec:background} recalls notation, robot modeling and a momentum-based control strategy for balancing with rigid, static contacts. Sections \ref{sec:Extension}--\ref{sec:control} detail the modeling and control framework designed for balancing in dynamic environments. Simulation results on humanoid robot iCub are presented in Section \ref{sec:results}. Conclusions and perspectives conclude the paper.

\section{Background}
\label{sec:background}

In this section, we provide a description of the modeling and control framework developed in \cite{Frontiers2015,pucciBalancing2016} for balancing a humanoid robot in a rigid environment. 

\subsection{Notation}
\begin{itemize}
\item  We denote with $\mathcal{I}$ the inertial frame of reference, whose $z$ axis points against the gravity, and with $\mathcal{S}$ a frame attached to the seesaw board, whose origin coincides with the seesaw center of mass.
\item  We make use of the subscript $s$ to distinguish the seesaw dynamic and kinematic quantities.
\item The constant $g$ denotes the norm of the gravitational acceleration. The constants $m$ and $m_s$ represent the masses of the robot and the seesaw, respectively. 
\item We denote with $S(x) \in \mathbb{R}^{3 \times 3}$ the skew-symmetric matrix such that $S(x)y = x \times y$, where $\times$ indicates the cross product operator in $\mathbb{R}^3$. 
\item Given a matrix $A \in \mathbb{R}^{m \times n}$, we denote with $A^{\dagger}\in \mathbb{R}^{n \times m}$ its Moore Penrose pseudoinverse. 
\item $e_i \in  \mathbb{R}^m$ is the canonical vector, consisting of all zeros but the $i$-th component that is equal to one.
\end{itemize}

\subsection{Modeling robot dynamics}

A robot is usually modeled as a set of $n + 1$ rigid bodies, namely \emph{links}, connected by $n$ joints with one degree of freedom each. We assume that the robot has no links with \emph{a priori} fixed position and orientation with respect to the inertial frame of reference, i.e. it is a \emph{free floating} system. 
Because of the above assumption, an element of the robot configuration space can be defined as $q = (\prescript{\mathcal{I}}{}p_{\mathcal{B}}, \prescript{\mathcal{I}}{}R_{\mathcal{B}}, q_j)$ and belongs to the Lie group $\mathbb{Q} = \mathbb{R}^3 \times SO{(3)} \times \mathbb{R}^n$. It is composed by the position and orientation of a \emph{base frame} $\mathcal{B}$ attached to a robot link w.r.t. the inertial reference frame, and the joints positions $q_j$. By differentiating $q$ we obtain the expression of system's velocities $\nu = ( ^\mathcal{I}\dot{ p}_{\mathcal{B}},^\mathcal{I}\omega_{\mathcal{B}},\dot{q}_j) = (\text{v}_{\mathcal{B}}, \dot{q}_j) \in \mathbb{R}^3 \times \mathbb{R}^3 \times \mathbb{R}^n$, where the angular velocity of the base frame $^\mathcal{I}\omega_{\mathcal{B}}$ is chosen such that $^\mathcal{I}\dot{R}_{\mathcal{B}} = S(^\mathcal{I}\omega_{\mathcal{B}})^\mathcal{I}{R}_{\mathcal{B}}$. 

It is assumed the robot is exerting $n_c$ distinct wrenches on the environment. Applying the Euler-Poincar\'e formalism as in \cite[Ch. 13.5]{Marsden2010} results in the following equations of motion:
\begin{align}
\label{eq:system}
{M}(q)\dot{{\nu}} + {C}(q, {\nu}){\nu} + {G}(q) =  B \tau + \sum_{k = 1}^{n_c} {J}^\top_{\mathcal{C}_k} f_k
\end{align}
where ${M} \in \mathbb{R}^{(n+6) \times (n+6)}$ is the mass matrix, ${C} \in \mathbb{R}^{(n+6) \times (n+6)}$ accounts for Coriolis and centrifugal effects, ${G} \in \mathbb{R}^{(n+6)}$ represents the gravity term, $B = (0_{n\times 6} , 1_n)^\top$ is a selector of the actuated degrees of freedom, $\tau \in \mathbb{R}^{n}$ is a vector representing the internal actuation torques, and $f_k \in \mathbb{R}^{6}$ denotes an external wrench applied by the environment to the link of the $k$-th contact. The Jacobian ${J}_{\mathcal{C}_k} = {J}_{\mathcal{C}_k}(q)$ is the map between the system's velocity ${\nu}$ and the linear and angular velocity at the $k$-th contact.
As described in \cite[Sec. 5]{traversaro2017}, it is possible to apply a coordinate transformation in the state space $(q,{\nu})$ that transforms the system dynamics~\eqref{eq:system} into a new form where
the mass matrix is block diagonal, thus decoupling joint and base frame accelerations. Also, in this new set of coordinates,  the first six rows of Eq. \eqref{eq:system} are the \emph{centroidal dynamics}\footnote{In the specialized literature, the term \emph{centroidal dynamics} 
is used to indicate the rate of change of the robot's momentum expressed at the center-of-mass, which then equals the summation of all external wrenches acting on the 
multi-body system \cite{Orin2013}.}. As an abuse of notation, we assume that system \eqref{eq:system} has been transformed into this new set of coordinates, i.e. 
\begin{IEEEeqnarray}{RCL}
\label{centrTrans}
M(q) &=& \begin{bmatrix} {M}_b(q) & 0_{6\times n} \\ 0_{n\times 6} & {M}_j(q) \end{bmatrix}, \quad
H    = M_b \text{v}_{\mathcal{C}},
\end{IEEEeqnarray}
with ${M}_b \in \mathbb{R}^{6\times 6}, {M}_j \in \mathbb{R}^{n\times n}$, ${H}:=(H_L^\top,H^\top_\omega)^\top\in \mathbb{R}^6$  the robot centroidal momentum, and $H_L, H_\omega \in \mathbb{R}^3$  the linear and angular momentum at the center of mass, respectively. The new base frame velocity is denoted by $\text{v}_{\mathcal{C}} \in \mathbb{R}^6$, which in the new coordinates yielding a block-diagonal mass matrix is given by $\text{v}_{\mathcal{C}} = (\dot{p}_c,\omega_o)$, where $\dot{p}_c \in \mathbb{R}^3$ is the velocity of the system's center of mass ${p}_c \in \mathbb{R}^3$, and $\omega_o \in \mathbb{R}^3$ is the so-called system's \emph{average angular velocity}.

Lastly, we assume that the location where a contact occurs on a link remains constant w.r.t. the inertial frame, i.e. the system is subject to a set of holonomic constraints of the form: ${J}_{\mathcal{C}_k}(q) {\nu} = 0.$
The constraints equations associated with all the rigid contacts can be represented as
\begin{IEEEeqnarray}{RCL}
\label{eqn:constraintsAll}
{J}(q) {\nu} = \begin{bmatrix}{J}_{\mathcal{C}_1}(q) \\ \cdots \\ {J}_{\mathcal{C}_{n_c}}(q)  \end{bmatrix}{\nu}  = J(q){\nu} = 0,
\IEEEeqnarraynumspace
\end{IEEEeqnarray}
where  $J = \begin{bmatrix} {J}_{\mathcal{C}_1}^\top & ... & {J}_{\mathcal{C}_{n_c}}^\top \end{bmatrix}^\top \in \mathbb{R}^{6n_c \times (n+6)}$ is the constraints jacobian.
By differentiating the kinematic constraint ~\eqref{eqn:constraintsAll}, one obtains
\begin{equation}
    \label{eq:constraints_acc}
   J\dot{\nu}+\dot{J}\nu = 0.
\end{equation}

In view of \eqref{eq:system}, the equations of motion along the constraints \eqref{eq:constraints_acc} are given by:
\begin{IEEEeqnarray}{LCL}
	\IEEEyesnumber
	{M}(q)\dot{{\nu}} + h(q,\nu) & = &  B \tau + J(q)^{\top} f \IEEEyessubnumber \label{eq:systemFinal}  \\
		   & s.t. \nonumber \\
    J\dot{\nu}+\dot{J}\nu = 0.\IEEEyessubnumber  
\end{IEEEeqnarray}
where $h:={C}(q, {\nu}){\nu} + {G}(q)  \in \mathbb{R}^{(n+6)}$, while $f=(f_1,\cdots,f_{n_c}) \in \mathbb{R}^{6n_c} $ are the set of contact wrenches -- i.e. Lagrange multipliers -- making Eq.~\eqref{eq:constraints_acc} satisfied. 

\subsection{Balancing control on static contacts}
\label{subsec:rigidControl}

We recall now the torque control strategy developed in previous publications \cite{Frontiers2015,pucciBalancing2016} for balancing a humanoid robot with static contacts. It is a \emph{task-based} control with two tasks: the task with higher priority is the control of robot centroidal momentum, while the secondary task is to stabilize the system's \emph{zero dynamics}. 

First, observe that the contact constraints equations Eq. \eqref{eq:constraints_acc} instantaneously relate the contact wrenches $f$ with the control input, namely the joint torques $\tau$. In fact, by substituting the state accelerations from Eq. \eqref{eq:systemFinal} into Eq. \eqref{eq:constraints_acc}, one has:
\begin{equation}
    \label{eq:relation-ft}
       J M^{-1}(J^{\top}f -h +B \tau) +\dot{J}\nu = 0.
\end{equation}
Writing explicitly the control torques from Eq. \eqref{eq:relation-ft} gives:
\begin{equation}
    \label{eq:torques}
    \tau = \Lambda^\dagger (J M^{-1}(h - J^\top f) - \dot{J}\nu) + N_\Lambda \tau_0
\end{equation}
with $\Lambda =  J M^{-1}B \in \mathbb{R}^{6n_c\times n}$;  $N_\Lambda \in \mathbb{R}^{n\times n}$  is the projector onto the nullspace of $\Lambda$,  and $\tau_0 \in \mathbb{R}^n$ is a free variable. 
If matrix $\Lambda$ has rank greater than the dimension of $f$, one can use $\tau$ to generate any desired contact wrenches $f^*$ by means of Eq. \eqref{eq:torques}.   

\vspace{4 mm}

\subsubsection{Momentum control}
 recall now that the rate-of-change of the robot momentum equals the net external wrenches acting on the robot, which in the present case reduces to the contact wrenches $f$ plus the gravity wrench:
\begin{IEEEeqnarray}{RCCCL}
\label{hDot}
 \dot{H}(f) &=& J_b^\top f - m g e_3,
\end{IEEEeqnarray}
where $H \in \mathbb{R}^6$ is the robot momentum, while because of the transformation applied in Eq. \eqref{centrTrans} matrix $J_b \in \mathbb{R}^{6n_c \times 6}$ can be obtained by partitioning the contact Jacobian $J = \begin{bmatrix} J_b & J_j \end{bmatrix}$. In view of Eq. \eqref{hDot}, and assuming that the contact wrenches can be chosen at will, we can choose $f$ such that:
\begin{IEEEeqnarray}{RCL}
    \label{hDotDes}
    \dot{H}(f) &=& 
    \dot{H}^* := \dot{H}^d - K_p \tilde{H} - K_i I_{\tilde{H}}    
\end{IEEEeqnarray}
where $H^d$ is the desired robot momentum, $\tilde{H} = H- H^d$ is the momentum error and $K_p, K_i {\in} \mathbb{R}^{6\times 6}$ two symmetric, positive definite matrices. The integral of robot momentum error, $I_{\tilde{H}}$ is obtained as described in \cite{nava2016}.

Observe that in case $n_c > 1$ (e.g. balancing on two feet), there are infinite sets of contact wrenches that satisfy Eq.~\eqref{hDotDes}. We parametrize the set of solutions $f^*$ as:
\begin{equation}
    \label{eq:forces}
    f^* = f_{H} + N_{b}f_0
\end{equation}
with $f_H =  (J_b^{\top})^{\dagger} \left(\dot{H}^* + m g e_3\right)$, $N_{b} \in \mathbb{R}^{6n_c \times 6n_c}$ the projector into the null space of $J_b^{\top}$, and $f_0\in \mathbb{R}^{6n_c}$ the wrench redundancy that does not influence $\dot{H}(f) = \dot{H}^*$. Then, the control torques that instantaneously realize the contact wrenches $f^*$ are given by Eq. \eqref{eq:torques}:
\begin{equation}
    \tau^* = \Lambda^\dagger (J M^{-1}(h - J^\top f^*) - \dot{J}\nu) + N_\Lambda \tau_0 \nonumber
\end{equation}

\subsubsection{Stability of the Zero Dynamics}
we are now left to define the free variable $\tau_0$, that may be used to ensure the stability of the so called \emph{zero dynamics} of the system \cite{ISIDORI2013}. A choice of $\tau_0$ that ensures the stability of the zero dynamics in case of one foot balancing is \cite{nava2016}:
\begin{IEEEeqnarray}{lCr}
\label{posturalNew}
    \tau_0 &=& h_j - J_j^\top f + u_0
\end{IEEEeqnarray}
where $u_0 := -K^j_{p}N_\Lambda M_j(q_j-q_j^d) -K^j_{d}N_\Lambda M_j\dot{q}_j$, and $K^{j}_p \in \mathbb{R}^{n \times n}$ and $K^{j}_d \in \mathbb{R}^{n \times n}$ two symmetric, positive definite matrices. An interesting property of the closed loop system~\eqref{eq:system}--\eqref{eq:torques}--\eqref{eq:forces}--\eqref{posturalNew} is recalled in the following Lemma.

\begin{lemma}[\cite{pucciTuning2016}]
\label{lemmaf0}
Assume that $\Lambda$ is full row rank. Then, the closed loop joint space dynamics does not depend upon the wrench redundancy $f_0$.
\end{lemma}

This result is a consequence of the postural control choice~\eqref{posturalNew} and it is of some interest: it means that the closed loop joint dynamics depends on the total rate-of-change of the momentum, i.e. $\dot{H}$, but not on the different forces generating it. Hence, any choice of the redundancy $f_0$ does not influence the joint dynamics, and we can exploit it to minimize the joint torques $\tau$ in Eq.~\eqref{eq:torques}.

In the language of \emph{Optimization Theory}, we can rewrite the  control strategy as the following optimization problem:
\begin{IEEEeqnarray}{RCL}
	\IEEEyesnumber
	\label{inputTorquesSOT}
	f^* &=& \argmin_{f}  |\tau^*(f)|^2 \IEEEyessubnumber \label{inputTorquesSOTMinTau}  \\
		   &s.t.& \nonumber \\
		   &&Af < b \IEEEyessubnumber  \label{frictionCones} \\
		   &&\dot{H}(f) = \dot{H}^*  \IEEEyessubnumber \\
		   &&\tau^*(f) = \argmin_{\tau}  |\tau - \tau_0(f)|^2 \IEEEyessubnumber	\label{optPost} 
  \\
		   	&& \quad s.t.  \nonumber \\
		   	&& \quad \quad \ \dot{J}(q,\nu)\nu + J(q)\dot{\nu} = 0
		    \IEEEyessubnumber 	\label{constraintsRigid} \\
		   	&& \quad \quad \ \dot{\nu} = M^{-1}(B\tau+J^\top f {-} h) \IEEEyessubnumber \\
		   && \quad \quad \ 	\tau_0 = 
		   h_j - J_j^\top f + u_0.		    \IEEEyessubnumber
\end{IEEEeqnarray}
The constraints~\eqref{frictionCones} ensure the satisfaction of friction cones, normal contact surface forces, and center-of-pressure constraints. The control torques are then given by $\tau {=} \tau^*(f^*)$.

\section{Modeling environment dynamics}
\label{sec:Extension}
The closed loop system \eqref{eq:system}--\eqref{eq:torques}--\eqref{eq:forces}--\eqref{posturalNew} exploits the assumption that the location where a contact occurs on a link remains constant w.r.t. the inertial frame, i.e. $J\nu = 0$. There are situations, however, in which the environment dynamics cannot be neglected. In this case, Eq. \eqref{eq:constraints_acc} becomes: 
\begin{equation}
   J\dot{\nu}+\dot{J}\nu = a_f. \nonumber
\end{equation}
where $a_f \in \mathbb{R}^{6n_c}$ represents the accelerations at the contact locations.
Our case study exemplifies this last situation: the robot is balancing with both feet leaning on a seesaw board of semi-cylindrical shape (Fig.\ref{fig:seesaw}).  
In what follows, we present a modeling and control framework derived from \eqref{eq:system}--\eqref{eq:torques}--\eqref{eq:forces}--\eqref{posturalNew} for two feet balancing on a seesaw.   

\subsection{Seesaw dynamics}

The seesaw can be considered a single rigid body with no \emph{a priori} fixed position and orientation w.r.t. the inertial frame. We also assume the seesaw is in contact with both the robot feet and a rigid ground, exerting on them the reaction forces and moments $-f$ and the contact forces and moments $f_s$, respectively. A complete description of the seesaw dynamics is given by the equations representing the rate of change of seesaw momentum, i.e. $\dot{H}_s$. In particular, we project the rate of change of seesaw momentum in the seesaw frame $\mathcal{S}$ previously defined, resulting in the following equations of motion: 
\begin{IEEEeqnarray}{RCL}
\label{eq:seesawEquations}
  M_s {^\mathcal{S}}\dot{\nu}_s + h_s &=& -J_r^\top f + J_s^\top f_s 
\end{IEEEeqnarray}
where $M_s \in \mathbb{R}^{6 \times 6}$ is the seesaw mass matrix, ${^\mathcal{S}}\dot{\nu}_s \in \mathbb{R}^6$ is the vector of seesaw linear and angular accelerations, $h_s \in \mathbb{R}^6$ represents the Coriolis and gravity terms. The jacobians $J_r$ and $J_s$ are the map between the seesaw velocity in the seesaw frame ${^\mathcal{S}}\nu_s$ and the velocities at the contacts locations. In particular, in the chosen representation the mass matrix $M_s$ is given by:
\begin{IEEEeqnarray}{RCL}
  M_s &=& \begin{bmatrix}
           m_s 1_3 & 0_3 \\
           0_3   &  {^\mathcal{S}}I_s
           \end{bmatrix} \nonumber
\end{IEEEeqnarray}
and matrix ${^\mathcal{S}}I_s \in \mathbb{R}^{3 \times 3}$ is constant, thus simplifying the formulation of seesaw dynamics. Further details on the derivation of Eq. \eqref{eq:seesawEquations} can be found in the Appendix \ref{App:1}. As an abuse of notation but for the sake of clarity let us omit from now on the superscript ${\mathcal{S}}$, e.g. ${^\mathcal{S}}\nu_s = \nu_s$. 

\subsection{Modeling contact constraints}

It is assumed that the robot feet are always attached to the seesaw, resulting in the following set of constraints:
\begin{IEEEeqnarray}{RCCCL}
\label{eqn:constraintsSeesawRobot}
\nu_{feet} &=& J_r \nu_s &=& J \nu.
\end{IEEEeqnarray}
Note that the above equation is coupling the seesaw and the robot dynamics. By differentiating Eq. \eqref{eqn:constraintsSeesawRobot}, one has:
\begin{equation}
    \label{eq:constraintsSeesawRobot_acc}
   J\dot{\nu}+\dot{J}\nu = J_r \dot{\nu}_s + \dot{J}_r \nu_s.
\end{equation}
Eq. \eqref{eq:constraintsSeesawRobot_acc} will substitute Eq. \eqref{eq:constraints_acc} in the formulation of the system's equations of motion. Lastly, we define the contact point $P$ as the intersection between the contact line of the seesaw with the ground and its frontal plane of symmetry. We assume that the seesaw is only rolling, and this implies that the linear velocity of the contact point $P$ is $v_p = 0$. Furthermore, let the seesaw frame be oriented as in figure \ref{fig:seesaw}. The shape of the (semi-cylindrical) seesaw constrains the rotation along the $y$ and $z$ axis. We model all the above mentioned constraints as follows:
\begin{IEEEeqnarray}{RCL}
\label{eqn:constraintsSeesawGround}
J_s \nu_s &=& 0,
\end{IEEEeqnarray}
and differentiating Eq. \eqref{eqn:constraintsSeesawGround} gives:
\begin{IEEEeqnarray}{RCL}
\label{eqn:constraintsSeesawGround_acc}
J_s \dot{\nu}_s + \dot{J}_s \nu_s &=& 0.
\end{IEEEeqnarray}
The shape of $J_r \in \mathbb{R}^{6n_c \times 6}$, $J_s \in \mathbb{R}^{5 \times 6}$ and their derivatives is described in the Appendix \ref{App:2}. Finally, the system dynamics for the robot balancing on a seesaw is given by the following set of equations:
\begin{IEEEeqnarray}{C}
\label{eqn:fullSystemDynamics}
\text{floating base dynamics}  \nonumber \\
M\dot{\nu} + h =  B \tau + J^\top f \IEEEyessubnumber	\label{dynRobot} \\
\text{seesaw dynamics}  \nonumber \\
M_s \dot{\nu}_s + h_s = -J_r^\top f + J_s^\top f_s \IEEEyessubnumber	\label{dynSeesaw} \\
\text{constraint: feet attached to the seesaw}  \nonumber \\
J\dot{\nu}+\dot{J}\nu = J_r \dot{\nu}_s + \dot{J}_r \nu_s \IEEEyessubnumber	\label{constrCoupling}  \\
\text{constraint: seesaw is rolling}  \nonumber \\
J_s \dot{\nu}_s + \dot{J}_s \nu_s = 0. \IEEEyessubnumber	\label{constrRolling} 
\end{IEEEeqnarray}
The above equations are valid for the specific case of a semi-cylindrical seesaw, but the approach we followed for obtaining Eq. \eqref{eqn:fullSystemDynamics} is more general, and can be reused in case the robot is interacting with different dynamic environments.

\section{The balancing control strategy}
\label{sec:control}

Given the effectiveness of control law \eqref{eq:torques}--\eqref{eq:forces}--\eqref{posturalNew} for balancing in a rigid environment, it may worth trying to extend this framework to the case of balancing on a seesaw. First, we make use of Eq. \eqref{eqn:constraintsSeesawGround_acc} to relate the feet wrenches $f$ with the contact forces and moments $f_s$. By substituting Eq. \eqref{eq:seesawEquations} into \eqref{eqn:constraintsSeesawGround_acc}, one has:
\begin{equation}
    \label{eq:relation-f-fs}
       J_s M_s^{-1}(J_s^{\top}f_s -h_s -J_r^\top f) +\dot{J}_s\nu_s = 0.
\end{equation}
Writing explicitly $f_s$ from Eq. \eqref{eq:relation-f-fs} gives:
\begin{equation}
    f_s = \Gamma^{-1}(J_s M_s^{-1}(h_s +J_r^\top f) -\dot{J}_s\nu_s), \nonumber
\end{equation}
with $\Gamma = (J_s M_s^{-1}J_s^{\top})$ . Now, substitute the above equation into Eq. \eqref{eq:seesawEquations}:
\begin{IEEEeqnarray}{C}
\label{eq:seesawEquationNofs}
  M_s \dot{\nu}_s + \bar{h}_s = A_s f 
\end{IEEEeqnarray}
where $\bar{h}_s = (1_6 -J_s^{\top}\Gamma ^{-1}J_s M_s^{-1})h_s + J_s^{\top}\dot{J}_s\nu_s$, while the matrix multiplying the feet forces and moments is given by $A_s = -(1_6 -J_s^{\top}\Gamma ^{-1}J_s M_s^{-1})J_r^\top$. It is worth noting that matrix $A_s$ is not full rank, but instead $\text{\emph{rank}}(A_s) = 1$. This is not surprising, in fact Eq. \eqref{eqn:constraintsSeesawGround_acc} implies the seesaw can only roll, and therefore it has only 1 degree of freedom left. Assuming the constraints Eq. \eqref{eqn:constraintsSeesawGround_acc} are always satisfied for any (reasonable) value of $f_s$, it means that the seesaw dynamics can be stabilized by controlling, for example, the seesaw angular momentum dynamics along the lateral direction, and this can be done by using the feet wrenches as a \emph{fictitious} control input of Eq. \eqref{eq:seesawEquationNofs}. Hence, one may think of controlling the whole system dynamics by means of the following control objectives:
\begin{itemize}
\item control of the robot momentum together with the seesaw angular momentum along the lateral direction by means of feet wrenches $f$;
\item ensure the stability of the system's zero dynamics as before by exploiting joint torques redundancy.
\end{itemize}
Concerning the primary control objective, the matrix projecting $f$ in the robot momentum and seesaw angular momentum equations can be obtained from Eq. \eqref{hDot} and Eq. \eqref{eq:seesawEquationNofs} and it is: $A_f = \begin{bmatrix} J_b & A_s^\top e_4 \end{bmatrix}^\top$. However, we performed a numerical analysis on matrix $A_f$ for different state configurations, by means of singular value decomposition (SVD) method. Numerical results point out that matrix $A_f$ is not full rank, but instead $\text{\emph{rank}}(A_s) = 6$. This implies that does not always exist a set $f^*$ of feet wrenches that can generate any desired trajectory for the given primary control task. 

\vspace{4 mm}

\subsubsection{Control of robot momentum only} being not possible to always control both the seesaw and the robot momentum dynamics, we first decide to control only the robot momentum as primary task, and then verify numerically that the seesaw angle trajectory still remains bounded within a limited range. For the given task, $f^*$ is the same of Eq. \eqref{eq:forces}:
\begin{equation}
    f^* = (J_b^{\top})^{\dagger} \left(\dot{H}^* + m g e_3\right) + N_{b}f_0 \nonumber
\end{equation}

\vspace{4 mm}

\subsubsection{Control of mixed momentum} another possibility is trying to control a quantity which depends on both the robot and the seesaw dynamics, for example the rate of change of system's momentum: 
\begin{IEEEeqnarray}{RCCCL}
\label{eq:totalMomentumDot}
  \dot{H}_t &=&  J_t^{\top} f_s -(m_s + m) g e_3 &=& J_t^{\top} A_t f + f_{bias} 
\end{IEEEeqnarray}
where Eq. \eqref{eq:totalMomentumDot} is obtained as described in Appendix \ref{App:3}. However, matrix $J_t \in \mathbb{R}^{5 \times 6}$: hence, the maximum rank of matrix $J_t^{\top} A_t$ is 5, and again it is not possible to always ensure the convergence of the total momentum to any desired trajectory by means of $f$. To overcome this problem, we decide to control the linear momentum of the robot only, together with the angular momentum of the whole system. The motivation behind this choice is not only to have a full rank matrix multiplying the feet wrenches, but it also exploits the difference of magnitude between the robot mass ($31 \text{Kg}$) and the seesaw mass ($4 \text{Kg}$), which implies the center of mass of the overall system is actually close to the center of mass of the robot. 

Consider now the following partitions of robot and total momentum:
\begin{IEEEeqnarray}{RCL}
    H := \begin{bmatrix}
         H_L \\
         H_\omega
         \end{bmatrix} , H_t := \begin{bmatrix}
                                H_{tL} \\
                                H_{t\omega}
                                \end{bmatrix} \nonumber
\end{IEEEeqnarray}
where $H_L, H_\omega \in \mathbb{R}^{3}$ are the linear and angular momentum of the robot, whereas $H_{tL}, H_{t\omega} \in \mathbb{R}^{3}$ are the system linear and angular momentum. We define the \emph{mixed} momentum as $H_m = \begin{bmatrix} H_L & H_{t\omega} \end{bmatrix}^\top$.
Being $\dot{H}_m = \dot{H}_m(f)$, we can choose $f$ such that:
\begin{IEEEeqnarray}{RCL}
    \label{hDotDesMixed}
    \dot{H}_m^* := \dot{H}_m^d - K_p \tilde{H}_m - K_i I_{\tilde{H}_m}    
\end{IEEEeqnarray}
where $H_m^d$ is the desired mixed momentum, $\tilde{H}_m = H_m- H_m^d$ is the momentum error and $K_p, K_i \in \mathbb{R}^{6\times 6}$ two symmetric, positive definite matrices. It is in general not possible to define an analytical expression for the integral of the angular momentum \cite{Orin2013}, and it is not straightforward to extend the results we presented in \cite{nava2016} to this new case. Therefore, $I_{\tilde{H}_m}$ is of the following form:
\begin{IEEEeqnarray}{RCL}
    I_{\tilde{H}_m} &=&   \begin{bmatrix}
                                m(x_{c}-x_{c}^d) \\
                                0_{3,1}
                          \end{bmatrix} \nonumber
\end{IEEEeqnarray}
where $x_c \in \mathbb{R}^3$ is the robot center of mass position. The feet wrenches $f^*$ that instantaneously realize the desired mixed momentum rate of change are then given by:
\begin{equation}
    f^* = A_m^{\dagger} \left(\dot{H}_m^* + \begin{bmatrix}
                                            S_L m g e_3 \\
                                           -S_\omega f_{bias} 
                                          \end{bmatrix}\right) + N_{m}f_0 \nonumber
\end{equation}
where $S_L = \begin{bmatrix} 1_3 & 0_3 \end{bmatrix}$, $S_\omega = \begin{bmatrix} 0_3 & 1_3 \end{bmatrix}$ are selector matrices; the multiplier of feet wrenches is $A_m =\begin{bmatrix} J_b S_L^{\top} &  A_t^{\top}  J_t S_\omega^{\top}  \end{bmatrix}^\top$ and $N_m$ is a null space projector.

\vspace{4 mm}

\subsubsection{Joint torques and zero dynamics} it is now possible to apply the same procedure presented in \ref{sec:background} for relating the joint torques $\tau$ with the desired feet wrenches $f^*$. By substituting the state accelerations from Eq. \eqref{dynRobot}--\eqref{eq:seesawEquationNofs} into Eq. \eqref{constrCoupling}, and writing explicitly the control torques one has:
\begin{equation}
    \label{eq:torquesSeesaw}
    \tau = \Lambda^\dagger (\bar{h} - \bar{J} f - \dot{J}\nu + \dot{J}_r\nu_s) + N_\Lambda \tau_0
\end{equation}
with $\bar{h} = JM^{-1}h - J_rM_s^{-1}h_s$, and $\bar{J} = JM^{-1}J^\top - J_rM_s^{-1}A_s $. The redundancy of joint torques is used for stabilizing the robot posture, and therefore $\tau_0$ remains the same of Eq. \eqref{posturalNew}. The wrench redundancy $f_0$ is still used for minimizing the joint torques. As before, we can rewrite the  control strategy as an optimization problem:
\begin{IEEEeqnarray}{RCL}
	\IEEEyesnumber
	\label{optSeesaw}
	f^* &=& \argmin_{f}  |\tau^*(f)|^2 \IEEEyessubnumber \label{inputTorquesSOTMinTauSeesaw}  \\
		   &s.t.& \nonumber \\
		   &&Af < b \IEEEyessubnumber \\
		   &&\dot{H}(f) = \dot{H}^*  \IEEEyessubnumber \label{seesawDotHOpt}\\
		   &&\tau^*(f) = \argmin_{\tau}  |\tau - \tau_0(f)|^2 \IEEEyessubnumber	\\
		   	&& \quad s.t.  \nonumber \\
		   	&& \quad \quad \ \dot{J}\nu + J\dot{\nu} = \dot{J}_r \nu_s + J_r \dot{\nu}_s \IEEEyessubnumber    \\
		   	&& \quad \quad \ \dot{J}_s \nu_s + J_s \dot{\nu}_s = 0 \IEEEyessubnumber 	\\
		   	&& \quad \quad \ \dot{\nu} = M^{-1}(B\tau+J^\top f {-} h) \IEEEyessubnumber \\
		   	&& \quad \quad \ \dot{\nu}_s = M_s^{-1}(J_s^\top f_s -J_r^\top f {-} h_s) \IEEEyessubnumber \\
		    && \quad \quad \ \tau_0 = h_j - J_j^\top f + u_0.		    \IEEEyessubnumber
\end{IEEEeqnarray}
where in case the primary control objective is the stabilization of the \emph{mixed} momentum trajectories, Eq. \eqref{seesawDotHOpt} is of the form: $\dot{H}_m(f) = \dot{H}_m^*$.

\section{Simulation results}
\label{sec:results}

We test the control solutions proposed in Section \ref{sec:Extension} by using a model of the humanoid robot iCub~\cite{Metta20101125} with 23 degrees-of-freedom (DoFs).  

\subsection{Simulation Environment}

Simulations are performed by means of a Simulink controller interfacing with Gazebo simulator~\cite{Koenig04}. The controller frequency is $100~\text{[Hz]}$. Of the different physic engines that can be used with Gazebo, we chose the Open Dynamics Engine (ODE). Furthermore, Gazebo integrates the dynamics with a fixed step semi-implicit Euler integration scheme. 
The advantage of using this simulation setup is twofold. First, we only have to specify the model of the robot, and the constraints arise naturally while simulating.
Another advantage of using Gazebo consists in the ability to test directly on the real robot the same control software used in simulation.

 \begin{figure}[t!]
   \centering
   \includegraphics[width=\columnwidth]{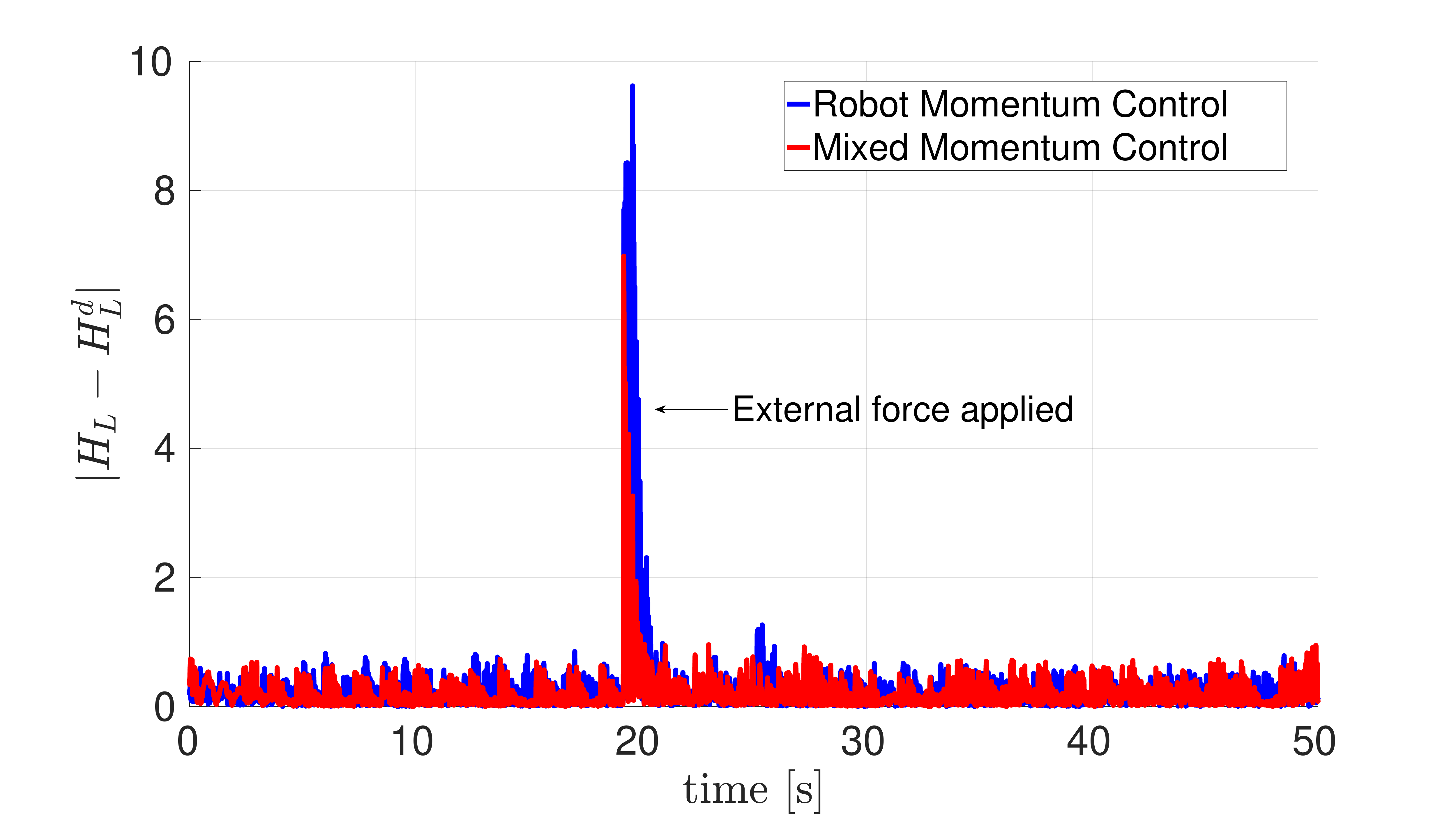}
   \caption{Norm of robot linear momentum error while balancing.}
   \label{fig:normR}
\end{figure}
 \begin{figure}[t!]
   \centering
   \includegraphics[width=\columnwidth]{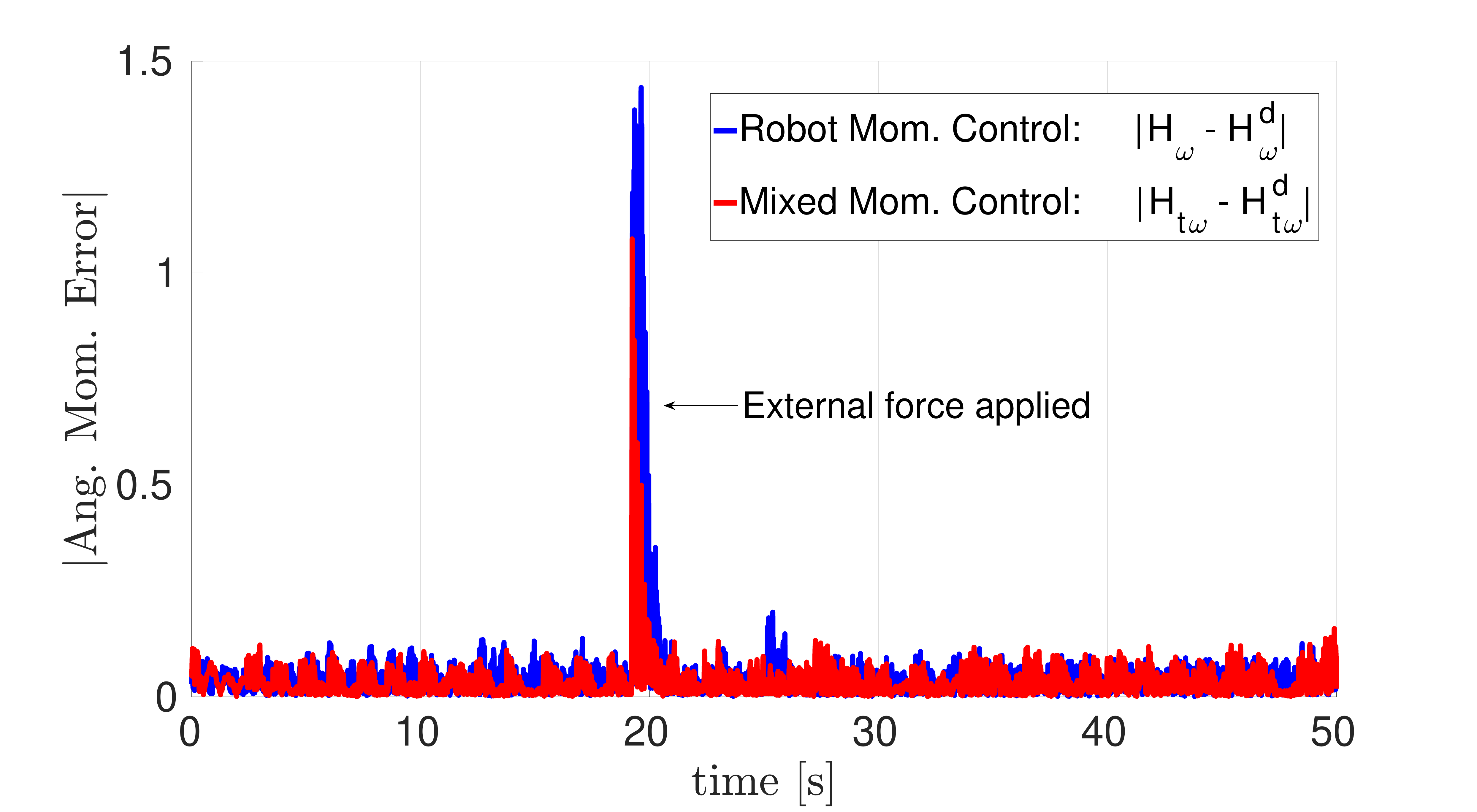}
   \caption{Norm of robot and system angular momentum error while balancing.}
   \label{fig:normM}
\end{figure}

\begin{figure}[t!]
   \centering
   \includegraphics[width=\columnwidth]{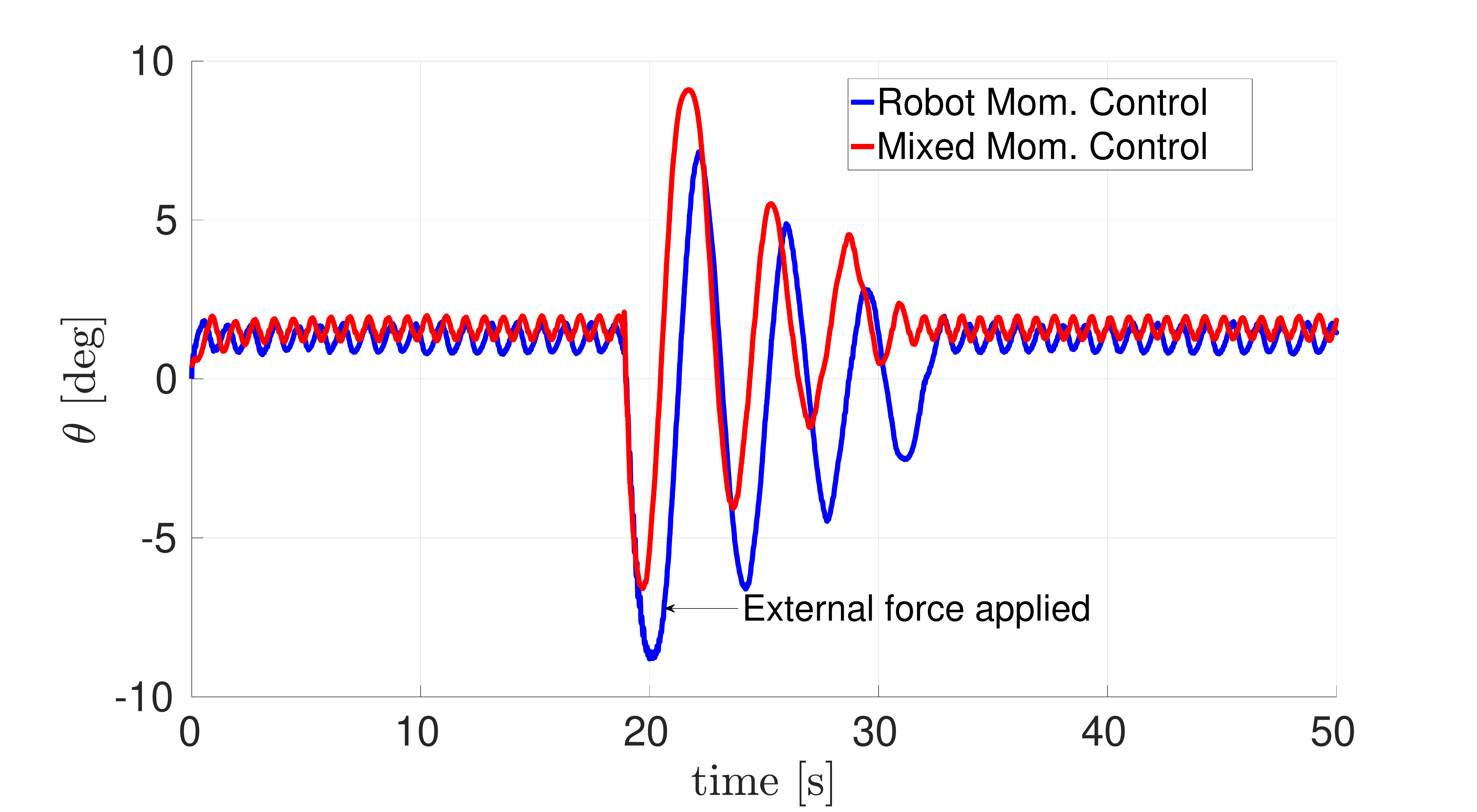}
   \caption{Seesaw orientation. Even if not explicitly controlled, its trajectory remains bounded while the robot is balancing.}
   \label{fig:theta}
\end{figure}

 \begin{figure}[t!]
   \centering
   \includegraphics[width=\columnwidth]{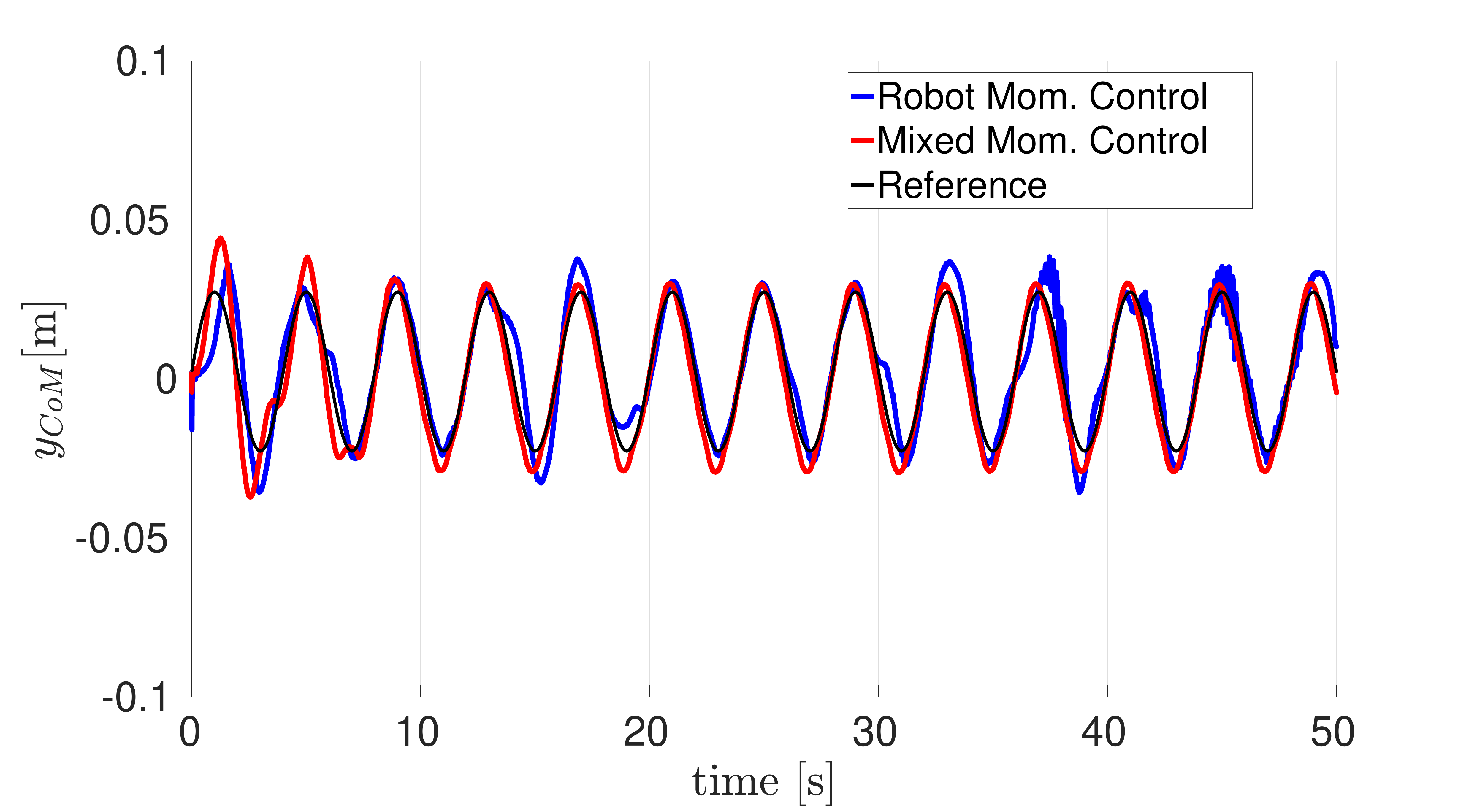}
   \caption{Lateral component of robot CoM position. Both controllers are able to track the desired CoM trajectory, but with mixed momentum control is possible to achieve better results.}
   \label{fig:com}
\end{figure}

\subsection{Robustness to external disturbances}

We first perform a robustness test on the closed loop system \eqref{eqn:fullSystemDynamics}--\eqref{optSeesaw}. The control objective is to stabilize the system about an equilibrium position. After $20~\text{[s]}$ an external force of amplitude $100~\text{[N]}$ is applied to the robot torso along the lateral direction for a period of $0.01~\text{[s]}$. Figure \ref{fig:normR} shows the norm of robot linear momentum error for both control laws. Analogously, Figure \ref{fig:normM} depicts the norm of robot and system angular momentum error when the primary task is to control the robot momentum and the mixed momentum, respectively. After the external force is applied, both controllers are still able to bring the system back to the equilibrium position. Figure \ref{fig:theta} shows instead the behaviour of the seesaw orientation $\theta$. The blue line represent $\theta$ when the control objective is to stabilize the robot momentum, while the red line is when the primary task is to control the mixed momentum. In both cases, the trajectory of $\theta$ remains bounded even after the application of the external force, and therefore both controllers are able to stabilize also the seesaw orientation, even if not explicitly controlled.

\subsection{Tracking Performances}

We then evaluate the two control laws for tracking a desired trajectory of the robot center of mass. The reference trajectory for the center of mass is a sinusoidal curve with amplitude of $2.5$ $[cm]$ and frequency of $0.25$ $[Hz]$ along the robot lateral direction. A dedicated gain tuning has been performed on both controllers in order to achieve better results. It is important to point out that the main scope of this analysis is not a comparison between the controllers performances, but rather to verify the stability of the closed loop system for the given task.
Figure \ref{fig:com} shows the lateral component of the robot center of mass position during the tracking. The black line is the reference trajectory. The red line is the center of mass position obtained with the mixed momentum control, and the blue line is obtained with the robot momentum control. Both controllers are able to track the desired trajectory. However, with the mixed momentum control is possible to achieve better results, while the other control strategy depicts poor tracking even after gain tuning. A possible explanation is that in order to keep balancing on the seesaw while tracking the CoM trajectory, the robot may be required to have an angular momentum reference different from zero. Trying to regulate the robot angular momentum to zero may worsen the tracking performances.

\section{Conclusions}
\label{sec:conclusions}

This paper proposes a modeling and control framework for balancing a humanoid robot in dynamic environments. In particular, the case study is the robot balancing on a seesaw board. The system equations of motion and the constraints are obtained following a general procedure, that can be reused in case the robot is in contact with a different object, or even a human. The control algorithm is redesigned taking into account the seesaw dynamics, and two different controllers are proposed. While both controllers show a good response to external disturbances, the mixed momentum control shows better tracking performances. 

In this paper, no experimental results are presented. However, preliminary tests on the real robot iCub have been performed, depicting some limitations of our control approach. For example, on real applications modeling errors and network delays can strongly affect the performances. Also, a proper estimation of parameters such as the seesaw orientation and angular velocity plays a very important role in the effectiveness of our control strategies. In order to be able to move on the real robot, future works might be focused on reducing the modeling and estimation errors, together with the design of a control law capable of dealing with model uncertainties, for example by means of adaptive control algorithms.

\section{Appendix}

\subsection{Seesaw dynamics in frame $\mathcal{S}$}
\label{App:1}

Let us define with $\mathcal{S}[\mathcal{I}]$ a reference frame whose origin is at the seesaw center of mass, and with the orientation of the inertial frame $\mathcal{I}$. The rate of change of seesaw momentum, when projected in this frame, is given by:
\begin{IEEEeqnarray}{RCL}
\label{eq:seesawMomentumWorld}
 ^{\mathcal{S}[\mathcal{I}]}\dot{H}_s &=&  -m_s g e_3 -\bar{J}_r^\top f + \bar{J}_s^\top f_s 
\end{IEEEeqnarray}
where the matrices $\bar{J}_r$ and $\bar{J}_s$ are defined in the next subsection \ref{App:2}.
Note that the mapping between frame $\mathcal{S}[\mathcal{I}]$ and the seesaw frame $\mathcal{S}$ is given by the relative rotation between the inertial frame and the seesaw frame, i.e. $^\mathcal{I} R_s $. Therefore, the projection of the seesaw momentum $^{\mathcal{S}[\mathcal{I}]}H_s$ into the seesaw frame $\mathcal{S}$ is given by: 
\begin{IEEEeqnarray}{RCCCL}
\label{eq:seesawMomentumTransf}
 ^{\mathcal{S}[\mathcal{I}]} H_s &=& ^\mathcal{I}\bar{R}_s {^\mathcal{S}}H_s &=&  \begin{bmatrix}
                                                                        ^\mathcal{I} R_s & 0_3 \\
                                                                        0_3 & ^\mathcal{I} R_s \end{bmatrix} {^\mathcal{S}}H_s.
\end{IEEEeqnarray}
By differentiating Eq. \eqref{eq:seesawMomentumTransf}, one has:
\begin{IEEEeqnarray}{RCL}
\label{eq:seesawMomentumTransfDot}
 ^{\mathcal{S}[\mathcal{I}]} \dot{H}_s &=& ^\mathcal{I}\dot{\bar{R}}_s {^\mathcal{S}}H_s + ^\mathcal{I}\bar{R}_s {^\mathcal{S}}\dot{H}_s.
\end{IEEEeqnarray}
Recall that ${^\mathcal{S}} H_s = M_s {^\mathcal{S}}\nu_s$, and that the derivative of a rotation matrix is given by: $^\mathcal{I}\dot{R}_s = {^\mathcal{I}}{R}_s S(^\mathcal{S} {\omega}_s)$, being  $^\mathcal{S} {\omega}_s$ the angular velocity of the seesaw projected in the seesaw frame. Also, in the seesaw frame the mass matrix $M_s$ is constant. Therefore, one has:
\begin{IEEEeqnarray}{RCL}
\label{eq:seesawMomentumTransfDotFinal}
 ^{\mathcal{S}[\mathcal{I}]} \dot{H}_s &=& ^\mathcal{I}{\bar{R}}_s ( \bar{S}(^\mathcal{S}{\omega}_s) M_s {^\mathcal{S}}\nu_s + M_s {^\mathcal{S}}\dot{\nu}_s).
\end{IEEEeqnarray}
with $\bar{S}(^\mathcal{S}{\omega}_s)$ a proper block diagonal matrix Finally, by substituting Eq. \eqref{eq:seesawMomentumTransfDotFinal} into \eqref{eq:seesawMomentumWorld}, and multiplying both sides by $^\mathcal{I}{\bar{R}^{-1}}_s = ^\mathcal{I}{\bar{R}^\top}_s$, one is left with Eq. \eqref{eq:seesawEquations}, where we define:
\begin{IEEEeqnarray}{L}
J_r =  \bar{J}_r {^\mathcal{I}}{\bar{R}}_s \nonumber \\ 
J_s =  \bar{J}_s {^\mathcal{I}}{\bar{R}}_s \nonumber \\ 
h_s =  \bar{S}(^\mathcal{S}{\omega}_s) M_s {^\mathcal{S}}\nu_s + ^\mathcal{I}{\bar{R}^\top}_s m_s g e_3 \nonumber 
\end{IEEEeqnarray}

\subsection{Derivation of matrices $J_r$ and $J_s$}
\label{App:2}
Recall the vector of feet linear and angular velocities expressed in the inertial frame:
\begin{IEEEeqnarray}{RCL}
{^\mathcal{I}} \nu_f &=& \begin{bmatrix}
                        {^\mathcal{I}}v_{lf}\\
                        {^\mathcal{I}}\omega_{lf}\\
                        {^\mathcal{I}}v_{rf}\\
                        {^\mathcal{I}}\omega_{rf}
                       \end{bmatrix}. \nonumber
\end{IEEEeqnarray}
The constraint of having the feet attached to the seesaw implies that ${^\mathcal{I}}\omega_{lf} = {^\mathcal{I}}\omega_{rf} = {^\mathcal{I}}\omega_{s}$. Also, one has:
\begin{IEEEeqnarray}{L}
{^\mathcal{I}}v_{lf} = {^\mathcal{I}}v_{s} -S({^\mathcal{I}}p_{sl}){^\mathcal{I}}\omega_{s} \nonumber \\
{^\mathcal{I}}v_{rf} = {^\mathcal{I}}v_{s} -S({^\mathcal{I}}p_{sr}){^\mathcal{I}}\omega_{s} \nonumber 
\end{IEEEeqnarray}
where ${^\mathcal{I}}p_{sl}$, ${^\mathcal{I}}p_{sr}$ represent the distance between the seesaw CoM and the left and right foot, respectively. Then,
\begin{IEEEeqnarray}{L}
{^\mathcal{I}} \nu_f = \begin{bmatrix}
                        1_3  & -S({^\mathcal{I}}p_{sl}) \\
                        0_3  & 1_3 \\
                        1_3  & -S({^\mathcal{I}}p_{sr}) \\
                        0_3  & 1_3 
                       \end{bmatrix}{^\mathcal{I}}\nu_{s} = \bar{J}_r {^\mathcal{I}}\nu_{s}\nonumber
\end{IEEEeqnarray}
Analogously, the constraint of only rolling implies that the velocity at the contact point $P$ between the seesaw and the ground is given by: ${^\mathcal{I}}v_{p} = {^\mathcal{I}}v_{s} -S({^\mathcal{I}}p_{sp}){^\mathcal{I}}\omega_{s} = 0$, thus implying ${^\mathcal{I}}v_{s} = S({^\mathcal{I}}p_{sp}){^\mathcal{I}}\omega_{s}$. The variable ${^\mathcal{I}}p_{sp}$ represents the distance between the seesaw CoM and the contact point $P$. Also, constraining the rotation along $y$ and $z$ axis implies the second and third component of the seesaw angular velocity are given by: $e_2^\top {^\mathcal{I}}\omega_{s} = e_3^\top {^\mathcal{I}}\omega_{s} = 0$. Then, one has:
\begin{IEEEeqnarray}{L}
\begin{bmatrix}
1_3  & -S({^\mathcal{I}}p_{sp}) \\
0_{1,3}  & e_2^\top \\
0_{1,3}  & e_3^\top 
\end{bmatrix}{^\mathcal{I}}\nu_{s} = \bar{J}_s {^\mathcal{I}}\nu_{s} = 0\nonumber
\end{IEEEeqnarray}
The matrices $J_s$ and $J_r$ are then obtained from $\bar{J}_s$ and $\bar{J}_r$ as described in the previous subsection. 

\subsection{Total momentum rate of change}
\label{App:3}

The system linear and angular momentum is obtained as a combination of the robot and seesaw momentum:
\begin{IEEEeqnarray}{RCL}
\label{eq:totalmomentum}
 {H}_t &=&  {^{t}X}_c^* H +  {^{t}X}_{\mathcal{S}[\mathcal{I}]}^* {^{\mathcal{S}[\mathcal{I}]}}H_s.
\end{IEEEeqnarray}
For not burdening the notation, we dropped the superscripts denoting the frames w.r.t. $H_t$ and $H$ are expressed. The transformation matrices in the space of wrenches $^t X_c^*$ and $^t X_{\mathcal{S}[\mathcal{I}]}^*$ are of the following form:
\begin{IEEEeqnarray}{L}
{^{t}X}_x^* = \begin{bmatrix}
1_3  & 0_3 \\
S({^\mathcal{I}}p_{x} - {^\mathcal{I}}p_{t})  & 1_3 \\
\end{bmatrix} \nonumber
\end{IEEEeqnarray}
where $^\mathcal{I}p_x = ^\mathcal{I}p_c$ for the robot momentum and $^\mathcal{I}p_x = ^\mathcal{I}p_{\mathcal{S}[\mathcal{I}]}$ for the seesaw momentum. Then, the derivative of Eq. \eqref{eq:totalmomentum} is given by:
\begin{IEEEeqnarray}{RCL}
 \dot{H}_t &=&  {^{t}\dot{X}}_c^* H +  {^{t}\dot{X}}_{\mathcal{S}[\mathcal{I}]}^* {^{\mathcal{S}[\mathcal{I}]}}H_s + {^{t}X}_c^* \dot{H} +  {^{t}X}_{\mathcal{S}[\mathcal{I}]}^* {^{\mathcal{S}[\mathcal{I}]}}\dot{H}_s. \nonumber 
\end{IEEEeqnarray}
Recall that the system's center of mass position is related to the robot and seesaw center of mass as follows: ${^\mathcal{I}}p_{t} = \frac{m {^\mathcal{I}}p_{c} + m_s {^\mathcal{I}}p_{{\mathcal{S}[\mathcal{I}]}}}{m + m_s}$. Also, recall that $S(x)x = 0$. Then, it is possible to verify that ${^{t}\dot{X}}_c^* H +  {^{t}\dot{X}}_{\mathcal{S}[\mathcal{I}]}^* {^{\mathcal{S}[\mathcal{I}]}}H_s = 0$, and therefore the rate of change of system's momentum is:
\begin{IEEEeqnarray}{RCL}
\label{eq:totalmomentumDer}
 \dot{H}_t &=& {^{t}X}_c^* \dot{H} +  {^{t}X}_{\mathcal{S}[\mathcal{I}]}^* {^{\mathcal{S}[\mathcal{I}]}}\dot{H}_s.
\end{IEEEeqnarray}
By substituting now Eq. \eqref{hDot},\eqref{eq:seesawMomentumWorld} into Eq. \eqref{eq:totalmomentumDer}, one has:
\begin{IEEEeqnarray}{RCL}
 \dot{H}_t &=& {^{t}X}_c^* (J_b^\top f - m g e_3) + \\ \nonumber
 & & {^{t}X}_{\mathcal{S}[\mathcal{I}]}^* (-m_s g e_3 -\bar{J}_r^\top f + \bar{J}_s^\top f_s). \nonumber
\end{IEEEeqnarray}
Observe that $J_b^\top$ and $\bar{J}_r^\top$, being the transformations mapping the wrenches from the contact locations to the seesaw and robot center of mass, are of the form: $J_b^\top = \begin{bmatrix} {^{c}X}^*_{l} & {^{c}X}^*_{r}\end{bmatrix}$ and $\bar{J}_r^\top = \begin{bmatrix} {^{\mathcal{S}[\mathcal{I}]}X}^*_{l} & {^{\mathcal{S}[\mathcal{I}]}X}^*_{r}\end{bmatrix}$. Thus $f$ is simplified from Eq. \eqref{eq:totalmomentumDer}. Furthermore, one can verify that $S({^\mathcal{I}}p_{\mathcal{S}[\mathcal{I}]} - {^\mathcal{I}}p_{t})m_s g e_3 +S({^\mathcal{I}}p_{c} - {^\mathcal{I}}p_{t})m g e_3  = 0$. This is consistent with the definition of $\dot{H}_t$ as the summation of all external wrenches, i.e the contact forces $f_s$ and the gravity wrench. By substituting $f_s$ into  Eq. \eqref{eq:totalmomentumDer} by means of \eqref{eq:relation-f-fs}, we finally obtain Eq. \eqref{eq:totalMomentumDot}, where we define: 
\begin{IEEEeqnarray}{LCL}
J_t^\top &= &{^{t}X}_{\mathcal{S}[\mathcal{I}]}^*\bar{J}_s^\top \nonumber \\ 
A_t &= & \Gamma^{-1}J_s M_s^{-1}J_r^\top  \nonumber \\ 
f_{bias} &=&  \Gamma^{-1}(J_s M_s^{-1}h_s -\dot{J}_s\nu_s) -(m_s+m) g e_3. \nonumber 
\end{IEEEeqnarray}

\addtolength{\textheight}{0cm}     

\bibliographystyle{IEEEtran}
\bibliography{IEEEabrv,Biblio}

\end{document}